\documentclass[11pt]{article}

\usepackage[final]{acl}

\usepackage{times}
\usepackage{latexsym}

\usepackage[T1]{fontenc}

\usepackage[utf8]{inputenc}

\usepackage{microtype}

\usepackage{inconsolata}

\usepackage{graphicx}

\usepackage{booktabs}
\usepackage{multirow}
\usepackage{graphicx}
\usepackage{array}
\usepackage{kotex}
\usepackage{colortbl}
\usepackage{amsmath}
\usepackage{amssymb}
\usepackage{enumitem}

\title{MixLID: Token-Level Language Identification in Mixed-Language Text}

\author{
    Seongtae Hong, Youngjoon Jang, Seungyoon Lee, Jungseob Lee, Heuiseok Lim\thanks{Corresponding authors} \\
    Department of Computer Science and Engineering, Korea University \\
    \texttt{ghdchlwls123@korea.ac.kr} \\
}

\begin{document}
\maketitle
\begin{abstract}
Processing multilingual text requires both identifying a broad range of languages and distinguishing the language of each token in inputs containing multiple languages. We introduce MixLID, a token-level language identification model supporting 111 language–script combinations. MixLID follows a two-stage training process: it first learns to distinguish individual languages using the language labels of monolingual examples, then continues training on mixed-language data. The second stage uses concatenated examples from existing code-switching data alongside augmented examples created through word substitution based on correspondences between English source words and loanwords. For monolingual identification, MixLID outperforms all baselines in both accuracy and macro F1 on four benchmarks within the language coverage shared with each baseline. For mixed-language identification, it outperforms the baselines in identifying the language of individual tokens on every evaluated language pair. MixLID achieves high performance across a broad range of languages in both monolingual and mixed-language settings. The model is publicly available at \url{https://huggingface.co/nlpai-lab/MixLID}.
\end{abstract}

\section{Introduction}
\label{sec:introduction}

Language identification (LID) determines the language used in a text, providing language information for constructing and using multilingual datasets~\citep{burchell2025expanded}. In large-scale corpus construction, LID is used to categorize text by language and filter for a target language~\citep{wenzek2020ccnet,culturax}. Errors in language identification can lead to the inclusion of text in other languages or the exclusion of target-language text, affecting the composition and quality of language-specific training data~\citep{penedo2025fineweb2,kreutzer2022quality}. Beyond data curation, LID is also used to evaluate whether generated text is in the requested language~\citep{chollampatt2025cross}. In reinforcement learning, language matching is incorporated into the reward to encourage models to respond in the target language~\citep{liu2026mending,zhang2026think}.

Treating an input as a single unit for language identification yields one language label for the entire text. Real-world text, however, may exhibit code-switching across sentences or within a sentence through the insertion of words or expressions from another language~\citep{elfardy2012token,burchell2024code}. A single label cannot distinguish the languages used in different spans of such an input~\citep{kargaran2024masklid}. This motivates token-level identification in addition to identifying the predominant language of the input~\citep{nguyen2013word,solorio2014overview}.

Prior work has expanded the language coverage of datasets and models to include low-resource languages~\citep{burchell2023open,kargaran2023glotlid,adebara2022afrolid}. For token-level LID in mixed-language inputs, supervised sequence labeling approaches use naturally occurring code-switched data with token-level language annotations to learn from word and character features and context~\citep{chittaranjan2014word,samih2016multilingual}. To address the scarcity of token-level annotations, synthetic data have also been constructed by randomly concatenating or mixing phrases from different languages and used for training alongside monolingual and naturally occurring code-switched data~\citep{zhang2018fast}. Training for mixed-language inputs thus requires considering both how expressions from different languages are combined and how monolingual and mixed-language data are used to train the model.

We introduce MixLID, a token-level language identification model supporting 111 language--script combinations.
The model first learns to distinguish individual languages using the language label of each monolingual example as supervision for its tokens. Training then continues on inputs formed by concatenating code-switched examples and inputs augmented through word substitutions based on English source words and their corresponding loanwords, to identify the language of individual tokens in mixed-language contexts. For monolingual identification, MixLID outperforms each baseline in accuracy and macro F1 on all four benchmarks, with each comparison restricted to the language--script combinations supported by both models. For mixed-language identification, it outperforms the baselines on every evaluated language pair. Analysis of the training stages shows that subsequent mixed-language training improves the identification of individual tokens while largely preserving performance on monolingual inputs.

\section{MixLID}
\label{sec:mixlid}
MixLID is trained in two stages to identify the language of each token in both monolingual and multilingual inputs. It first learns to distinguish languages from monolingual examples in a multilingual corpus, then continues training on existing and synthetic examples containing multiple languages.

\subsection{Training Data Construction}
\paragraph{Monolingual Data}
\label{sec:monolingual-supervision}
We use monolingual examples with sentence-level language labels to construct token-level training data. Specifically, we use OpenLID-v3~\citep{fedorova2026openlid} data covering 111 language--script combinations and assign each sentence's language label to its constituent words. However, the language label of a sentence may not be appropriate for all expressions it contains. We therefore assign separate labels to numbers, punctuation, and symbols. Abbreviations found in the lists\footnote{We use the abbreviation lists provided by NLTK Punkt.} receive a language label that covers the entire expression, including periods. Abbreviations in the English list receive the English label regardless of the source language, whereas those found only in the source-language list retain the source-language label. We further apply a heuristic filter based on English letter patterns to examples in non-Latin scripts to reduce incorrect assignment of the source-language label to expressions from other languages, while retaining these abbreviations.

\paragraph{Mixed-Language Data}
\label{sec:mixed-language-supervision}

To identify the language of each token in mixed-language contexts, we use both code-switched data and data augmented through word substitution. We concatenate examples from the ACS Corpus into longer training sequences while preserving their token-level language labels. For augmentation, we apply word substitutions to sentences and documents from Smol~\citep{smol} and passages from Belebele~\citep{belebele}, retaining each augmented text as a separate training example. We construct the substitution dictionary from Wiktionary etymological data extracted with Wiktextract~\citep{wiktextract}, retaining direct borrowing relationships between English source words and loanwords in other languages. We apply these dictionary mappings to each text independently, replacing English source words with corresponding loanwords or vice versa. Each replacement is assigned the label of the selected target language, while the remaining expressions are assigned the language label of the original text by default. All data and substitution candidates are restricted to supported languages. We retain the monolingual labeling rules for abbreviations, numbers, punctuation, and symbols, but omit the exclusion rules based on Latin character patterns to preserve mixed-language inputs. Detailed preprocessing and labeling rules for both data types are provided in Appendix~\ref{sec:training-details}.

\begin{table*}[t!]
    \centering
    \setlength{\tabcolsep}{2.2pt}
    \setlength{\aboverulesep}{0pt}
    \setlength{\belowrulesep}{0pt}
    \renewcommand{\arraystretch}{1.05}
    \resizebox{\textwidth}{!}{%
    \begin{tabular}{@{}ll*{4}{>{\columncolor{black!6}}c>{\columncolor{black!6}}ccc}@{}}
    \toprule
    \multirow{2}{*}{Benchmark} & \multirow{2}{*}{Metric} & \multicolumn{2}{>{\columncolor{black!6}}c}{OpenLID-v3} & \multicolumn{2}{c}{GlotLID-v3} & \multicolumn{2}{>{\columncolor{black!6}}c}{NLLB-LID} & \multicolumn{2}{c}{fastText lid.176} & \multicolumn{2}{>{\columncolor{black!6}}c}{CLD3} & \multicolumn{2}{c}{CommonLingua} & \multicolumn{2}{>{\columncolor{black!6}}c}{CONLID} & \multicolumn{2}{c}{Polyglot Tagger} \\
    \cmidrule(lr){3-4}\cmidrule(lr){5-6}\cmidrule(lr){7-8}\cmidrule(lr){9-10}\cmidrule(lr){11-12}\cmidrule(lr){13-14}\cmidrule(lr){15-16}\cmidrule(lr){17-18}
     & & Baseline & \textbf{MixLID} & Baseline & \textbf{MixLID} & Baseline & \textbf{MixLID} & Baseline & \textbf{MixLID} & Baseline & \textbf{MixLID} & Baseline & \textbf{MixLID} & Baseline & \textbf{MixLID} & Baseline & \textbf{MixLID} \\
    \midrule
    \multirow{3}{*}{FLORES+} & \#L & \multicolumn{2}{>{\columncolor{black!6}}c}{110} & \multicolumn{2}{c}{110} & \multicolumn{2}{>{\columncolor{black!6}}c}{106} & \multicolumn{2}{c}{70} & \multicolumn{2}{>{\columncolor{black!6}}c}{74} & \multicolumn{2}{c}{96} & \multicolumn{2}{>{\columncolor{black!6}}c}{110} & \multicolumn{2}{c}{70} \\
     & Acc & 99.19 & \textbf{99.77} & 99.38 & \textbf{99.77} & 98.97 & \textbf{99.75} & 95.59 & \textbf{99.94} & 97.37 & \textbf{99.91} & 95.40 & \textbf{99.89} & 98.89 & \textbf{99.77} & 99.46 & \textbf{99.91} \\
     & Macro F1 & 99.42 & \textbf{99.76} & 99.57 & \textbf{99.76} & 99.15 & \textbf{99.75} & 95.47 & \textbf{99.94} & 97.31 & \textbf{99.91} & 96.18 & \textbf{99.89} & 99.32 & \textbf{99.76} & 99.58 & \textbf{99.92} \\
    \midrule
    \multirow{3}{*}{WiLI-2018} & \#L & \multicolumn{2}{>{\columncolor{black!6}}c}{82} & \multicolumn{2}{c}{82} & \multicolumn{2}{>{\columncolor{black!6}}c}{80} & \multicolumn{2}{c}{70} & \multicolumn{2}{>{\columncolor{black!6}}c}{70} & \multicolumn{2}{c}{81} & \multicolumn{2}{>{\columncolor{black!6}}c}{82} & \multicolumn{2}{c}{67} \\
     & Acc & 97.58 & \textbf{98.43} & 97.72 & \textbf{98.41} & 97.52 & \textbf{98.37} & 95.88 & \textbf{98.91} & 95.76 & \textbf{98.59} & 97.05 & \textbf{98.41} & 97.64 & \textbf{98.41} & 98.58 & \textbf{98.84} \\
     & Macro F1 & 98.16 & \textbf{98.58} & 98.21 & \textbf{98.58} & 97.98 & \textbf{98.54} & 96.34 & \textbf{98.98} & 96.34 & \textbf{98.71} & 97.82 & \textbf{98.58} & 98.20 & \textbf{98.58} & 98.88 & \textbf{98.92} \\
    \midrule
    \multirow{3}{*}{UDHR} & \#L & \multicolumn{2}{>{\columncolor{black!6}}c}{101} & \multicolumn{2}{c}{101} & \multicolumn{2}{>{\columncolor{black!6}}c}{98} & \multicolumn{2}{c}{67} & \multicolumn{2}{>{\columncolor{black!6}}c}{72} & \multicolumn{2}{c}{91} & \multicolumn{2}{>{\columncolor{black!6}}c}{101} & \multicolumn{2}{c}{67} \\
     & Acc & 98.82 & \textbf{99.38} & 98.14 & \textbf{99.38} & 98.36 & \textbf{99.34} & 94.99 & \textbf{99.87} & 97.60 & \textbf{99.84} & 93.45 & \textbf{99.80} & 97.10 & \textbf{99.38} & 99.33 & \textbf{99.84} \\
     & Macro F1 & 98.75 & \textbf{99.32} & 99.14 & \textbf{99.32} & 98.40 & \textbf{99.29} & 95.30 & \textbf{99.86} & 97.23 & \textbf{99.84} & 94.63 & \textbf{99.81} & 98.36 & \textbf{99.32} & 99.54 & \textbf{99.84} \\
    \midrule
    \multirow{3}{*}{CommonLID} & \#L & \multicolumn{2}{>{\columncolor{black!6}}c}{58} & \multicolumn{2}{c}{58} & \multicolumn{2}{>{\columncolor{black!6}}c}{54} & \multicolumn{2}{c}{39} & \multicolumn{2}{>{\columncolor{black!6}}c}{45} & \multicolumn{2}{c}{53} & \multicolumn{2}{>{\columncolor{black!6}}c}{58} & \multicolumn{2}{c}{41} \\
     & Acc & 94.04 & \textbf{96.95} & 94.56 & \textbf{96.75} & 93.93 & \textbf{96.66} & 95.93 & \textbf{97.74} & 90.28 & \textbf{97.25} & 87.07 & \textbf{96.81} & 90.43 & \textbf{96.75} & 95.42 & \textbf{97.74} \\
     & Macro F1 & 81.85 & \textbf{85.79} & 83.47 & \textbf{85.80} & 83.28 & \textbf{86.28} & 84.96 & \textbf{90.17} & 75.72 & \textbf{88.51} & 78.30 & \textbf{86.50} & 78.56 & \textbf{85.80} & 85.07 & \textbf{88.51} \\
    \bottomrule
    \end{tabular}%
    }
    \caption{Sentence-level language identification performance (\%) on the language--script intersection between MixLID and each baseline. \#L denotes the intersection size. Bold indicates the higher score in each paired comparison.}
    \label{tab:sentence-lid-pairwise}
\end{table*}

\subsection{Learning Objective}
\label{sec:token-level-identification}
We align the labels in the constructed data with subword tokens using character offsets in the input text. When a word is split into multiple subwords, all resulting subwords are assigned the same language label as the original word. A multilingual encoder produces a contextual representation for each subword, and a classification head predicts its label. Let $x=(x_1,\ldots,x_T)$ denote the input subword sequence and $y_i\in\mathcal{Y}$ the label associated with each labeled subword $x_i$. The set $\mathcal{I}$ contains the positions of labeled subwords used for training. The label set $\mathcal{Y}$ includes the supported language labels and separate labels for numbers, punctuation, and symbols, and is shared across both training stages. Let $p_{\theta}(c\mid x,i)$ denote the predicted probability of label $c\in\mathcal{Y}$ at position $i$, where $\theta$ comprises the parameters of the encoder and classification head. We train the model to predict the corresponding token labels by minimizing the cross-entropy loss:
\begin{equation}
\mathcal{L}(\theta;x,y)
=-\frac{1}{|\mathcal{I}|}
\sum_{i\in\mathcal{I}}
\log p_{\theta}(y_i\mid x,i).
\label{eq:token-loss-en}
\end{equation}
The same objective is used in both stages, with training on mixed-language data continuing from the model trained on monolingual data.

\begin{table}[t]
    \centering
    \small
    \resizebox{\columnwidth}{!}{%
    \begin{tabular}{llllcr}
    \toprule
    Training & Dataset & Source & Augmentation & \#L & \#Samples \\
    \midrule
    Mono & OpenLID-v3 & Mono & --- & 111 & 123.4M \\
    \midrule
    \multirow{4}{*}{Mixed}
    & ACS Corpus & CS & concatenation & 10 & 509.8K \\
    & SmolSent & Mono & Substitution & 87 & 46.8K \\
    & SmolDoc & Mono & Substitution & 92 & 20.7K \\
    & Belebele & Mono & Substitution & 111 & 56.6K \\
    \bottomrule
    \end{tabular}%
    }
    \caption{Training data composition and statistics.}
    \label{tab:training-data}
    \end{table}

\section{Experiments}
\subsection{Setup}
\paragraph{Training}

We use mmBERT-small~\citep{marone2025mmbertmodernmultilingualencoder} as the backbone of MixLID with a 114-class token classification head for 111 language--script combinations and three auxiliary labels (\texttt{NUM}, \texttt{PUNCT}, and \texttt{SYM}). Table~\ref{tab:training-data} summarizes the training data. Training proceeds in two stages, beginning with approximately 123.4M monolingual examples from OpenLID-v3. The resulting model is then further trained on approximately 634K examples constructed from the ACS Corpus, Smol, and Belebele for mixed-language identification. ACS examples are shuffled and concatenated in groups of up to four. For Smol and Belebele, loanword-based substitution is applied independently to each eligible expression with a probability of 70\%. Detailed training settings for both stages and hardware specifications are provided in the Appendix~\ref{sec:training-details}.

\paragraph{Evaluation}

For sentence-level evaluation, we compare MixLID with OpenLID-v3~\citep{fedorova2026openlid}, GlotLID-v3~\citep{kargaran2023glotlid}, NLLB-LID~\citep{nllb}, fastText lid.176~\citep{fasttext}, CLD3\footnote{\url{https://github.com/google/cld3}}, CommonLingua~\citep{commonlingua}, CONLID~\citep{foroutan2026conlid}, and Polyglot Tagger\footnote{\url{https://huggingface.co/polyglot-tagger/language-identification}} on four benchmarks: FLORES+~\citep{flores}, WiLI-2018~\citep{thoma2018wili}, UDHR~\citep{kargaran2023glotlid}\footnote{\url{https://huggingface.co/datasets/cis-lmu/udhr-lid}}, and CommonLID~\citep{suarez2026commonlid}. Each comparison is restricted to languages jointly supported by MixLID and the baseline. MixLID and Polyglot Tagger determine the input language by majority voting over token predictions. We report overall accuracy and macro F1, the mean of per-language F1 scores. For token-level evaluation, we compare against MaskLID~\citep{kargaran2024masklid}, Polyglot Tagger, and CLD3 on ACS~\citep{sterner2025minimal} and the LID subset of CodeMixBench~\citep{yang2025codemixbench}. CLD3 provides language predictions over text spans, which we align with the reference tokens. We retain language combinations supported by four models and evaluate the same eligible tokens. Token accuracy and macro F1 are computed per example and averaged within each language combination.

\subsection{Results}
\begin{figure*}[t!]
    \centering
    \includegraphics[width=0.95\textwidth]{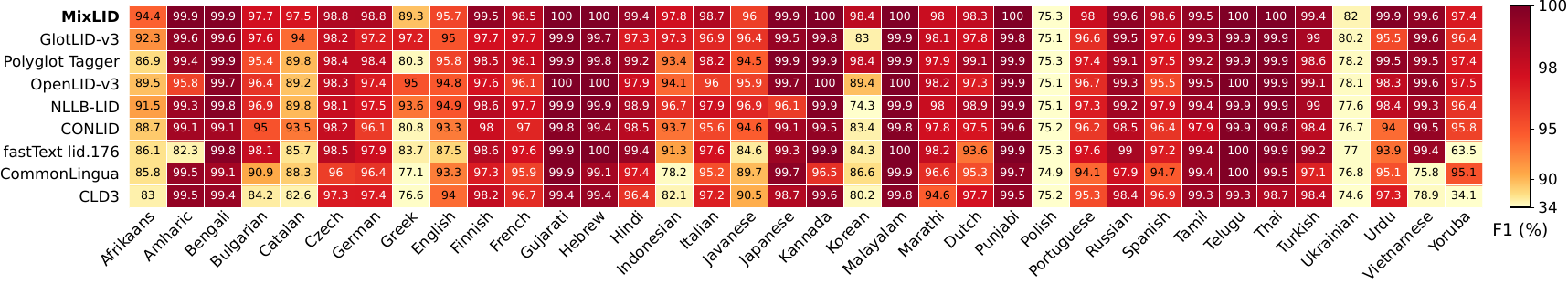}
    \caption{Language-wise F1 scores for the 36 languages shared by all models and all four sentence-level benchmarks. Each cell shows the average score across these datasets.}
    \label{fig:language-wise-performance}
\end{figure*}

\paragraph{Monolingual LID at the Sentence Level}
Table~\ref{tab:sentence-lid-pairwise} shows that MixLID outperforms all eight baselines in both accuracy and macro F1 across the four benchmarks. This advantage holds over both sentence-level identification models and Polyglot Tagger, which supports token-level identification. Although MixLID is trained to predict language labels for individual tokens, these results show that its predictions also support accurate identification of the predominant language of an input. The higher accuracy and macro F1 indicate an advantage in both overall prediction accuracy and performance averaged across languages. The results demonstrate consistent performance advantages across evaluation data from diverse sources, including web text.

\begin{table}[t]
    \centering
    \footnotesize
    \setlength{\tabcolsep}{4.5pt}
    \resizebox{0.9\columnwidth}{!}{%
    \begin{tabular}{@{}llc@{\hspace{15pt}}c@{\hspace{15pt}}c@{\hspace{15pt}}c@{}}
    \toprule
    & Pair & CLD3 & MaskLID & Polyglot & \textbf{MixLID} \\
    \midrule
    \multirow{12}{*}{ACS} & da--en & 56.1 / 32.6 & 38.7 / 20.7 & 62.5 / 37.9 & \textbf{83.3} / \textbf{73.5} \\
     & de--en & 56.4 / 33.9 & 35.9 / 22.9 & 78.2 / 62.0 & \textbf{98.0} / \textbf{96.9} \\
     & es--en & 48.2 / 28.2 & 22.9 / 15.3 & 74.6 / 56.9 & \textbf{96.1} / \textbf{94.1} \\
     & fr--en & 54.8 / 32.1 & 24.4 / 15.5 & 71.0 / 47.7 & \textbf{93.1} / \textbf{90.4} \\
     & id--en & 31.7 / 18.4 & 19.3 / 12.1 & 38.0 / 24.2 & \textbf{92.1} / \textbf{89.1} \\
     & it--en & 56.3 / 33.3 & 29.6 / 19.3 & 74.3 / 55.7 & \textbf{95.6} / \textbf{93.6} \\
     & nl--en & 52.2 / 30.9 & 35.8 / 21.9 & 66.1 / 45.1 & \textbf{92.6} / \textbf{90.0} \\
     & sv--en & 56.6 / 32.9 & 41.7 / 24.9 & 71.7 / 48.7 & \textbf{92.3} / \textbf{87.1} \\
     & tr--de & 58.1 / 35.1 & 53.1 / 37.2 & 86.6 / 80.5 & \textbf{93.9} / \textbf{81.5} \\
     & tr--en & 44.6 / 25.8 & 31.2 / 18.9 & 69.9 / 45.6 & \textbf{91.6} / \textbf{87.4} \\
     & zh--en & 86.2 / 69.5 & 37.5 / 25.2 & 94.7 / 89.2 & \textbf{97.0} / \textbf{93.0} \\
    \cmidrule(l){2-6}
     & \textbf{Avg} & 54.6 / 33.9 & 33.6 / 21.3 & 71.6 / 53.9 & \textbf{93.2} / \textbf{88.8} \\
    \midrule
    \multirow{4}{*}{CodeMixBench} & de--en & 86.3 / 61.1 & 55.3 / 29.2 & 90.6 / 65.8 & \textbf{95.2} / \textbf{82.8} \\
     & es--en & 55.9 / 47.3 & 25.8 / 17.6 & 82.3 / 71.2 & \textbf{92.8} / \textbf{88.7} \\
     & zh--en & 80.5 / 46.0 & 61.6 / 31.0 & 83.9 / 55.3 & \textbf{95.5} / \textbf{90.3} \\
    \cmidrule(l){2-6}
     & \textbf{Avg} & 74.2 / 51.5 & 47.6 / 25.9 & 85.6 / 64.1 & \textbf{94.5} / \textbf{87.3} \\
    \bottomrule
    \end{tabular}%
    }
    \caption{Token-level LID results (Acc / Macro F1).}
    \label{tab:token-lid}
\end{table}

\paragraph{Mixed-Language LID at the Token Level}
Table~\ref{tab:token-lid} shows that MixLID outperforms MaskLID, Polyglot Tagger, and CLD3 in both accuracy and macro F1 across all evaluated subsets of ACS and CodeMixBench. Averaged across language pairs, MixLID achieves an accuracy of 93.23\% and a macro F1 of 88.78\% on ACS, and 94.50\% and 87.29\%, respectively, on CodeMixBench. These results indicate that MixLID can distinguish expressions from different languages within the same context, beyond identifying the predominant language of an input. Its advantage across the evaluated language combinations suggests that this ability is not limited to a particular language pair.

\section{Analysis}
\label{sec:analysis}

\subsection{Language-wise Performance}
\label{sec:language-wise-performance}

To examine whether the overall performance advantage of MixLID extends to individual languages, we compare performance on the 36 languages shared by all models and all four benchmarks, as shown in Figure~\ref{fig:language-wise-performance}. MixLID achieves the highest F1 on 26 of these languages. Its mean F1 is 97.38\%, exceeding the strongest baseline, GlotLID-v3, by 0.99 percentage points. It outperforms GlotLID-v3 and Polyglot Tagger on 33 and 31 languages, respectively, and achieves F1 scores above 95\% on 32 of the 36 languages. These results indicate that the overall performance advantage of MixLID is reflected across a broad range of languages.

\begin{table}[t]
\centering
\setlength{\tabcolsep}{2.3pt}
\renewcommand{\arraystretch}{1.0}
\resizebox{0.87\columnwidth}{!}{%
\begin{tabular}{l*{6}{>{\centering\arraybackslash}p{3.9em}}}
\toprule
 & \multicolumn{2}{c}{Stage 1} & \multicolumn{2}{c}{Stage 2} & \multicolumn{2}{c}{Stage 1+2} \\
\cmidrule(lr){2-3}\cmidrule(lr){4-5}\cmidrule(lr){6-7}
Benchmark & Acc & Macro F1 & Acc & Macro F1 & Acc & Macro F1 \\
\midrule
FLORES+   & 99.67 & 99.64 & 97.39 & 97.21 & \textbf{99.77} & \textbf{99.76} \\
WiLI-2018 & \textbf{98.52} & \textbf{98.62} & 94.05 & 94.74 & 98.43 & 98.58 \\
UDHR      & 98.69 & 98.34 & 96.32 & 95.86 & \textbf{99.38} & \textbf{99.32} \\
CommonLID & 93.64 & 81.42 & 90.91 & 74.67 & \textbf{96.95} & \textbf{85.79} \\
\midrule
ACS          & 53.75 & 34.31 & 87.32 & 80.78 & \textbf{93.23} & \textbf{88.78} \\
CodeMixBench & 77.36 & 54.41 & 78.55 & 69.65 & \textbf{94.50} & \textbf{87.29} \\
\bottomrule
\end{tabular}%
}
\caption{Effect of training stages on LID performance at the sentence and token levels across six benchmarks.}
\label{tab:training-ablation}
\end{table}

\subsection{Effect of Training Stages}
\label{sec:Stages}

Table~\ref{tab:training-ablation} compares training on monolingual data alone (Stage 1), mixed-language data alone (Stage 2), and both stages in sequence (Stage 1+2). Stage1 outperforms Stage2 on all four sentence-level benchmarks, whereas Stage2 performs better at token identification in mixed-language inputs. The model trained through both stages outperforms either individual stage on the mixed-language benchmarks ACS and CodeMixBench. It also surpasses Stage1 in sentence-level identification, except for a slight decrease on WiLI-2018. These results suggest that the two stages serve complementary roles, improving mixed-language identification while largely preserving monolingual performance.

\section{Conclusion}
We introduced MixLID, a token-level language identification model covering 111 language labels and designed to distinguish languages within mixed-language inputs. Trained on monolingual data followed by existing and augmented mixed-language data, MixLID outperformed the baselines in both monolingual and mixed-language identification. Our analysis showed that mixed-language training improved individual token identification while largely preserving performance on monolingual inputs. We hope MixLID will serve as a useful resource for research and applications requiring language identification within multilingual text.

\section*{Limitations}
The linguistic scope of this study is limited to the 111 language labels supported by MixLID and does not encompass the full range of linguistic diversity. In addition, dictionary-based substitution and automatic labeling may not fully account for contextual usage and linguistic ambiguity, potentially introducing unnatural expressions or labeling errors into the training data. Finally, mixed-language evaluation covers a limited set of code-switching language combinations, and generalization to a wider range of combinations requires further validation.

\section*{Ethical Considerations}
This study uses publicly available datasets and lexical resources to develop a model for language identification in text. The loanword-based augmented data are constructed for training token-level language identification and do not represent the actual language use or code-switching practices of any particular linguistic community. To ensure transparency in data use and experimentation, we document the resource sources, data construction and preprocessing methods, and evaluation procedures.



\bibliography{custom}
\appendix
\section{Related Work}

Research on language identification at the sentence and document levels has explored statistical classification using character and word features as well as neural approaches~\citep{jauhiainen2019automatic}. More recent work has expanded language coverage through the collection, curation, and release of training data, addressing diverse scripts and low-resource languages~\citep{burchell2023open,kargaran2023glotlid,adebara2022afrolid}. Efforts to better distinguish closely related languages and handle nonlinguistic noise have included enriching training data, merging selected language varieties, and treating noise as a separate class~\citep{fedorova2026openlid}. Evaluations on web data have shown that high scores on established benchmarks do not necessarily translate into comparable performance in domains containing noise and diverse language varieties~\citep{suarez2026commonlid}. A systematic comparison of factors including input length, training data size, and domain has likewise identified cross-domain generalization as a major challenge~\citep{van2025identifying}. To address domain differences, supervised contrastive learning has been used to learn language representations and improve identification of low-resource languages on out-of-domain data~\citep{foroutan2026conlid}.

Word-level identification in mixed-language text has been formulated as sequence labeling that accounts for context, with weak supervision derived from monolingual data~\citep{king2013labeling}. Synthetic mixed-language data have also been used to train token classifiers supporting multiple languages~\citep{zhang2018fast}. Research has explored identification using lexicons, morphological analysis, and contextual embeddings, as well as comparisons between encoder-based classification and decoder-based generation~\citep{sterner2023tongueswitcher,iranzo2025word}. Other work has harmonized annotations across language-pair datasets to train token-level identification of English versus other languages and examined transfer to language pairs unseen during training~\citep{sterner2024multilingual}. Token identification models have also been fine-tuned on authentic code-switched data and synthetic examples constructed by inserting contiguous spans from another language into sentences~\citep{signoroni2026rebelot}. An openly available model similarly fine-tunes a pretrained encoder to identify the language of each token using synthetic data formed by concatenating sentences in different languages. Other approaches predict the set of languages present in a mixed-language input through sentence-level multilabel classification, or iteratively mask language-specific features to detect additional languages with an existing identifier~\citep{burchell2024code,kargaran2024masklid}. Our work focuses on two-stage training from monolingual to mixed-language data, using both existing code-switched data and augmented data based on loanword correspondences for mixed-language training. This approach supports token-level language identification in mixed-language inputs across a broad range of languages, while aggregating the predictions also enables identification of the language of a monolingual input.

\begin{table*}[t!]
    \centering
    \resizebox{\textwidth}{!}{%
    \begin{tabular}{*{4}{>{\raggedright\arraybackslash}p{10em}>{\centering\arraybackslash}p{6em}}}
    \toprule
    Language & Label & Language & Label & Language & Label & Language & Label \\
    \midrule
    Afrikaans & \texttt{afr\_Latn} & Hausa & \texttt{hau\_Latn} & Marathi & \texttt{mar\_Deva} & Spanish & \texttt{spa\_Latn} \\
    Amharic & \texttt{amh\_Ethi} & Hebrew & \texttt{heb\_Hebr} & Modern Greek & \texttt{ell\_Grek} & Standard Estonian & \texttt{ekk\_Latn} \\
    Arabic & \texttt{ara\_Arab} & Hindi & \texttt{hin\_Deva} & Nepali & \texttt{npi\_Deva} & Standard Latvian & \texttt{lvs\_Latn} \\
    Armenian & \texttt{hye\_Armn} & Hungarian & \texttt{hun\_Latn} & Nigerian Fulfulde & \texttt{fuv\_Latn} & Standard Malay & \texttt{zsm\_Latn} \\
    Assamese & \texttt{asm\_Beng} & Icelandic & \texttt{isl\_Latn} & North Azerbaijani & \texttt{azj\_Latn} & Sundanese & \texttt{sun\_Latn} \\
    Bambara & \texttt{bam\_Latn} & Igbo & \texttt{ibo\_Latn} & Northern Uzbek & \texttt{uzn\_Latn} & Swahili & \texttt{swh\_Latn} \\
    Basque & \texttt{eus\_Latn} & Iloko & \texttt{ilo\_Latn} & Norwegian Bokmål & \texttt{nob\_Latn} & Swati & \texttt{ssw\_Latn} \\
    Bengali & \texttt{ben\_Beng} & Indonesian & \texttt{ind\_Latn} & Nyanja & \texttt{nya\_Latn} & Swedish & \texttt{swe\_Latn} \\
    Bulgarian & \texttt{bul\_Cyrl} & Italian & \texttt{ita\_Latn} & Odia & \texttt{ory\_Orya} & Tajik & \texttt{tgk\_Cyrl} \\
    Burmese & \texttt{mya\_Mymr} & Japanese & \texttt{jpn\_Jpan} & Panjabi & \texttt{pan\_Guru} & Tamil & \texttt{tam\_Taml} \\
    Catalan & \texttt{cat\_Latn} & Javanese & \texttt{jav\_Latn} & Paraguayan Guaraní & \texttt{gug\_Latn} & Telugu & \texttt{tel\_Telu} \\
    Cebuano & \texttt{ceb\_Latn} & Kabuverdianu & \texttt{kea\_Latn} & Pedi & \texttt{nso\_Latn} & Thai & \texttt{tha\_Thai} \\
    Central Khmer & \texttt{khm\_Khmr} & Kachin & \texttt{kac\_Latn} & Persian & \texttt{fas\_Arab} & Tibetan & \texttt{bod\_Tibt} \\
    Central Kurdish & \texttt{ckb\_Arab} & Kannada & \texttt{kan\_Knda} & Plateau Malagasy & \texttt{plt\_Latn} & Tigrinya & \texttt{tir\_Ethi} \\
    Croatian & \texttt{hrv\_Latn} & Kazakh & \texttt{kaz\_Cyrl} & Polish & \texttt{pol\_Latn} & Tosk Albanian & \texttt{als\_Latn} \\
    Czech & \texttt{ces\_Latn} & Kinyarwanda & \texttt{kin\_Latn} & Portuguese & \texttt{por\_Latn} & Tsonga & \texttt{tso\_Latn} \\
    Danish & \texttt{dan\_Latn} & Kirghiz & \texttt{kir\_Cyrl} & Romanian & \texttt{ron\_Latn} & Tswana & \texttt{tsn\_Latn} \\
    Dutch & \texttt{nld\_Latn} & Korean & \texttt{kor\_Hang} & Russian & \texttt{rus\_Cyrl} & Turkish & \texttt{tur\_Latn} \\
    English & \texttt{eng\_Latn} & Lao & \texttt{lao\_Laoo} & Serbian & \texttt{srp\_Cyrl} & Ukrainian & \texttt{ukr\_Cyrl} \\
    Filipino & \texttt{fil\_Latn} & Lingala & \texttt{lin\_Latn} & Shan & \texttt{shn\_Mymr} & Urdu & \texttt{urd\_Arab} \\
    Finnish & \texttt{fin\_Latn} & Lithuanian & \texttt{lit\_Latn} & Shona & \texttt{sna\_Latn} & Vietnamese & \texttt{vie\_Latn} \\
    French & \texttt{fra\_Latn} & Luo & \texttt{luo\_Latn} & Sindhi & \texttt{snd\_Arab} & Waray & \texttt{war\_Latn} \\
    Ganda & \texttt{lug\_Latn} & Macedonian & \texttt{mkd\_Cyrl} & Sinhala & \texttt{sin\_Sinh} & West Central Oromo & \texttt{gaz\_Latn} \\
    Georgian & \texttt{kat\_Geor} & Malayalam & \texttt{mal\_Mlym} & Slovak & \texttt{slk\_Latn} & Wolof & \texttt{wol\_Latn} \\
    German & \texttt{deu\_Latn} & Maltese & \texttt{mlt\_Latn} & Slovenian & \texttt{slv\_Latn} & Xhosa & \texttt{xho\_Latn} \\
    Gujarati & \texttt{guj\_Gujr} & Mandarin Chinese & \texttt{cmn\_Hans} & Somali & \texttt{som\_Latn} & Yoruba & \texttt{yor\_Latn} \\
    Haitian & \texttt{hat\_Latn} & Mandarin Chinese & \texttt{cmn\_Hant} & Southern Pashto & \texttt{pbt\_Arab} & Zulu & \texttt{zul\_Latn} \\
    Halh Mongolian & \texttt{khk\_Cyrl} & Maori & \texttt{mri\_Latn} & Southern Sotho & \texttt{sot\_Latn} &  &  \\
    \bottomrule
    \end{tabular}%
    }
    \caption{Supported languages and their training labels.}
    \label{tab:supported-languages}
    \end{table*}
    
\section{Training Details}
\label{sec:training-details}

\subsection{Language Inventory}
\label{sec:language-inventory}

Table~\ref{tab:supported-languages} lists the languages supported by MixLID and the 111 language labels used for training. We add three labels, \texttt{NUM}, \texttt{PUNCT}, and \texttt{SYM}, for numbers, punctuation, and symbols, respectively, yielding 114 output classes. These labels are assigned to the corresponding expressions instead of a language label. The same label inventory is used in both training stages.

\subsection{Shared Labeling and Token Alignment}
\label{sec:shared-labeling}

We distinguish numbers, punctuation, and symbols using Unicode categories, while treating decimal points and commas separating groups of three digits as part of a numeric expression. For example, \texttt{3.14} and \texttt{1,000} each receive a single \texttt{NUM} label, whereas \texttt{3,14} is split into \texttt{NUM}, \texttt{PUNCT}, and \texttt{NUM}. Percent and per mille signs and their variants are labeled \texttt{SYM} regardless of their Unicode categories. Abbreviations matching the NLTK Punkt lists receive a language label over the entire expression, including periods. Matches in the English list take precedence and receive the English label; those found only in the list for the original language retain that language label.

Apostrophes within words receive the same language label as the word, whereas standalone quotation marks are labeled \texttt{PUNCT}. Ordinary spaces are left unlabeled, while newlines and tabs in mixed-language data are labeled \texttt{PUNCT}. The resulting labeled spans are aligned with subwords using character offsets. When a subword overlaps multiple spans, we assign the label of the span with the greatest character overlap, excluding ordinary spaces. Ties are resolved in favor of the earlier span.

\subsection{Monolingual Data Preparation}
\label{sec:monolingual-preprocessing}

To reduce the assignment of a sentence's language label to expressions from other languages, we apply a filter based on English letter patterns to monolingual examples in non-Latin scripts. Except for expressions in the abbreviation lists, examples containing a sequence of three or more English letters are excluded. This filter is not applied to examples in Latin scripts or to mixed-language data.

\subsection{Mixed-Language Data Preparation}
\label{sec:mixed-preprocessing}

\paragraph{Code-Switched Data Preparation}

To train token-level language identification on longer inputs containing multiple languages, we construct training sequences by concatenating existing code-switched examples. We use the ACS Corpus, which provides token-level language labels. Before concatenation, we remove anonymized user mentions and relabel tokens without a specific language label as numbers, punctuation, or symbols. Examples containing any remaining tokens that cannot be relabeled are excluded. We shuffle the cleaned examples and concatenate up to four with a single space between them, preserving their language labels and using each example once.

\paragraph{Loanword-Based Augmentation}

To construct code-switched data, we apply loanword-based substitution to datasets containing monolingual examples in multiple languages: sentences and documents from Smol and passages from Belebele. The substitution dictionary includes entries for borrowing, adapted borrowing, unadapted borrowing, learned borrowing, semi-learned borrowing, and orthographic borrowing. We exclude entries whose forms are identical to their English source words or differ only in case, as well as affix forms with a leading or trailing hyphen. For reverse substitution, we also exclude loanwords associated with multiple English source words. Within each input, we select the longest non-overlapping matches at word boundaries and independently substitute each match with a probability of 70\%. When an English source word has candidates in multiple languages, we select a candidate language uniformly at random. We do not perform contextual word sense disambiguation or infer inflected forms, and we retain examples even when no substitution is applied.

\begin{table}[t]
\centering
\small
\begin{tabular}{lcc}
\toprule
Setting & Monolingual & Mixed \\
\midrule
Optimizer & AdamW & AdamW \\
LR schedule & Linear & Linear \\
Maximum length & 8,192 & 8,192 \\
Learning rate & $5\times10^{-5}$ & $1\times10^{-5}$ \\
Epochs & 1 & 1 \\
Batch size per GPU & 512 & 128 \\
Gradient accumulation & 2 & 1 \\
Effective batch size & 4,096 & 512 \\
Warmup ratio & 0.05 & 0.05 \\
Weight decay & 0 & 0 \\
Precision & BF16 & BF16 \\
Seed & 42 & 42 \\
\bottomrule
\end{tabular}
\caption{Training hyperparameters for both stages.}
\label{tab:training-hyperparameters}
\end{table}

\subsection{Training Configuration}
\label{sec:optimization}
Table~\ref{tab:training-hyperparameters} reports the hyperparameters for each training stage. The second stage is initialized from the model trained in the first stage, with the optimizer and learning rate scheduler reinitialized. Both stages update all model parameters using four NVIDIA RTX A6000 GPUs.

\begin{table*}[t]
\centering
\footnotesize
\resizebox{0.85\textwidth}{!}{%
\begin{tabular}{@{}>{\centering\arraybackslash}p{0.13\textwidth}p{0.16\textwidth}p{0.65\textwidth}@{}}
\toprule
Type & Name & URL \\
\midrule
Model & OpenLID-v3 & \url{https://huggingface.co/HPLT/OpenLID-v3} \\
 & GlotLID-v3 & \url{https://huggingface.co/cis-lmu/glotlid} \\
 & NLLB-LID & \url{https://huggingface.co/facebook/fasttext-language-identification} \\
 & fastText lid.176 & \url{https://fasttext.cc/docs/en/language-identification.html} \\
 & CommonLingua & \url{https://huggingface.co/PleIAs/CommonLingua} \\
 & CONLID & \url{https://huggingface.co/epfl-nlp/ConLID} \\
 & CLD3 & \url{https://github.com/google/cld3} \\
 & MaskLID & \url{https://github.com/cisnlp/MaskLID} \\
 & Polyglot Tagger & \url{https://huggingface.co/polyglot-tagger/language-identification} \\
\midrule
Benchmark & FLORES+ & \url{https://huggingface.co/datasets/openlanguagedata/flores_plus} \\
 & WiLI-2018 & \url{https://doi.org/10.5281/zenodo.841984} \\
 & UDHR & \url{https://huggingface.co/datasets/cis-lmu/udhr-lid} \\
 & CommonLID & \url{https://huggingface.co/datasets/commoncrawl/CommonLID} \\
 & ACS Benchmark & \url{https://huggingface.co/datasets/igorsterner/acs-benchmark} \\
 & CodeMixBench & \url{https://huggingface.co/datasets/CodeMixBench/CodeMixBench} \\
\bottomrule
\end{tabular}%
}
\caption{Sources of baseline models and evaluation benchmarks.}
\label{tab:evaluation-sources}
\end{table*}
\section{Evaluation Details}
\label{sec:evaluation-details}

\subsection{Baseline Models}
\label{sec:baseline-models}
The baselines include models that predict a single language for an entire input and models that provide token-level or span-level predictions. The coverage figures below refer to the label inventories of the models used in our experiments and are distinct from the numbers of shared evaluation languages reported in the main tables.

\begin{itemize}[leftmargin=*, labelsep=0.4em]
\item \textbf{OpenLID-v3} A fastText-based model trained on monolingual data from multiple sources, with a focus on improving identification precision for closely related languages. It provides 194 language labels and a non-language label.

\item \textbf{GlotLID-v3} A fastText-based model designed to extend language identification coverage to low-resource languages. It provides 2,102 labels distinguishing languages and scripts.

\item \textbf{NLLB-LID} A fastText-based language identification model used for data collection and filtering in multilingual machine translation. It predicts the language of an entire input, and the model used in our experiments provides 218 output labels.

\item \textbf{fastText lid.176} A general-purpose language identification model trained on Wikipedia, Tatoeba, and SETimes. It supports 176 languages and predicts the language of an input text.

\item \textbf{CommonLingua} A lightweight byte-level model designed for language identification in large corpora. Its training data include diverse documents and low-resource languages, and it provides 334 language labels.

\item \textbf{CONLID} A language identification model that uses contrastive learning and supports a broad range of languages. The model used in our experiments provides 2,099 output labels.

\item \textbf{CLD3} A neural language identification model using character $n$-gram features, with 109 language and script labels. It provides predictions for both entire inputs and text spans and is therefore included in both evaluations.

\item \textbf{MaskLID} A method that applies iterative masking to an existing language identification model to identify expressions in different languages within mixed-language inputs, without additional training. We use GlotLID-v3, with its 2,102 labels, as the underlying model.

\item \textbf{Polyglot Tagger} An XLM-R-based token classification model trained on synthetic data constructed by combining sentences in different languages. The checkpoint used in our experiments supports 105 languages. We use its token predictions directly for token-level evaluation and aggregate them for sentence-level evaluation.
\end{itemize}

\subsection{Evaluation Benchmarks}
\label{sec:evaluation-benchmarks}

\paragraph{Monolingual Benchmarks}

We use four benchmarks to evaluate the prediction of a single language for each input. FLORES+ contains parallel texts translated into multiple languages; we use the \texttt{devtest} split of v4.5. WiLI-2018 consists of paragraphs collected from Wikipedia, for which we use the official test split. UDHR is based on multilingual translations of the Universal Declaration of Human Rights; we use the cleaned version released with the GlotLID study. CommonLID contains web text with human-annotated language labels. Each sentence, paragraph, or line is treated as one input, and each comparison is restricted to the languages jointly supported by the models.

\paragraph{Mixed-Language Benchmarks}
We use the ACS Benchmark and the LID subset of CodeMixBench to evaluate token-level language identification in mixed-language inputs. ACS provides observed code-switched sentences and manipulated variants. We evaluate both types of input across 11 language pairs in its test split, using the provided automatic language labels. For CodeMixBench, we evaluate German--English, Spanish--English, and Chinese--English, which are supported by all models in the comparison.

\subsection{Evaluation Protocol and Metrics}
\label{sec:evaluation-protocol}

\paragraph{Language Matching}
We map language codes to a common representation to reconcile differences in model label conventions. Sentence-level comparisons include only examples in languages supported by the benchmark and both models. Arabic and Persian varieties are mapped to the corresponding macrolanguage labels when these are provided by a model; support for one variety is not taken to imply support for other varieties. Script constraints are also applied when specified by the model. We use the script metadata provided by FLORES+ and UDHR, and infer the predominant script from the distribution of Unicode scripts in the input for WiLI-2018 and CommonLID. For fastText lid.176, CommonLingua, and Polyglot Tagger, we apply language-level coverage and the script constraints of the comparison model. Token-level evaluation includes only subsets for which MixLID, MaskLID, Polyglot Tagger, and CLD3 all support both languages in the pair.

\paragraph{Prediction Alignment}
For sentence-level evaluation, MixLID and Polyglot Tagger determine the input language by majority voting over subword language predictions, excluding non-language labels. Ties are resolved by lexicographic label order. For token-level evaluation, we use character offsets in the original text to align model output spans with the reference tokens. For each reference token, we sum the character overlap with predicted spans by language and select the language with the largest total overlap. CLD3 span offsets are converted from bytes to characters, and tokens with no corresponding prediction are treated as undetermined. We exclude tokens with non-language, mixed, or ambiguous reference labels, as well as tokens consisting entirely of whitespace, numbers, punctuation, or symbols, consistently across models. Undetermined predictions and predictions outside the language pair are counted as errors for eligible tokens.

\paragraph{Sentence-Level Metrics}
We report accuracy and macro F1 to measure overall identification accuracy and performance across languages. Accuracy is the proportion of evaluation examples for which the language is correctly predicted.
\begin{equation}
\operatorname{Acc}=\frac{\text{\# correct samples}}{\text{\# samples}}
\end{equation}
For each language $\ell$, F1 is computed from its true positives, false positives, and false negatives. Let $\mathcal{L}$ denote the set of shared languages included in the evaluation.
\begin{equation}
F_{1,\ell}=\frac{2\operatorname{TP}_{\ell}}{2\operatorname{TP}_{\ell}+\operatorname{FP}_{\ell}+\operatorname{FN}_{\ell}}
\end{equation}
\begin{equation}
\operatorname{MacroF1}=\frac{1}{|\mathcal{L}|}\sum_{\ell\in\mathcal{L}}F_{1,\ell}
\end{equation}
Thus, accuracy measures the proportion of correct predictions across all evaluation examples, whereas macro F1 is the arithmetic mean of the F1 scores for individual languages.

\paragraph{Token-Level Metrics}
For mixed-language inputs, we first compute accuracy and macro F1 over the eligible tokens in each example, then average the scores across examples within each language pair. Examples with no eligible tokens are excluded. Let $A_i$ and $F_i$ denote the accuracy and macro F1 for example $i$, respectively, and let $\mathcal{L}_i$ be the union of its reference and predicted labels over eligible tokens.

\begin{equation}
A_i=\frac{\text{\# correct tokens}}{\text{\# tokens}}
\end{equation}
\begin{equation}
F_i=\frac{1}{|\mathcal{L}_i|}\sum_{\ell\in\mathcal{L}_i}F_{1,\ell}^{(i)}
\end{equation}
Here, $F_{1,\ell}^{(i)}$ is computed using only the eligible tokens in example $i$. Avg in the main tables is the arithmetic mean of the language-pair scores within each benchmark. All scores are reported as percentages.

\end{document}